# AutoCOR: Autonomous Condylar Offset Ratio Calculator on TKA-Postoperative Lateral Knee X-ray

Gulsade Rabia Cakmak, Ibrahim Ethem Hamamci, Mehmet Kursat Yilmaz, Reda Alhajj, Ibrahim Azboy, Mehmet Kemal Ozdemir

**Abstract— The postoperative range of motion is one of the crucial factors indicating the outcome of Total Knee Arthroplasty (TKA). Although the correlation between range of knee flexion and posterior condylar offset (PCO) is controversial in the literature, PCO maintains its importance on evaluation of TKA. Due to limitations on PCO measurement, two novel parameters, posterior condylar offset ratio (PCOR) and anterior condylar offset ratio (ACOR), were introduced. Nowadays, the calculation of PCOR and ACOR on plain lateral radiographs is done manually by orthopedic surgeons. In this regard, we developed a software, AutoCOR, to calculate PCOR and ACOR autonomously, utilizing unsupervised machine learning algorithm (k-means clustering) and digital image processing techniques. The software AutoCOR is capable of detecting the anterior/posterior edge points and anterior/posterior cortex of the femoral shaft on true postoperative lateral conventional radiographs. To test the algorithm, 50 postoperative true lateral radiographs from Istanbul Kosuyolu Medipol Hospital Database were used (32 patients). The mean PCOR was 0.984 (SD 0.235) in software results and 0.972 (SD 0.164) in ground truth values. It shows strong and significant correlation between software and ground truth values (Pearson r=0.845 p<0.0001). The mean ACOR was 0.107 (SD 0.092) in software results and 0.107 (SD 0.070) in ground truth values. It shows moderate and significant correlation between software and ground truth values (Spearman's rs=0.519 p=0.0001412). We suggest that AutoCOR is a useful tool that can be used in clinical practice.**

**_Index Terms—_ TKA, Image Processing, K-means Clustering, Posterior Condylar Offset, Anterior Condylar Offset**

## I. INTRODUCTION

Total Knee Arthroplasty (TKA) is a surgical operation applied to the patients who are in a severe stage of osteoarthritis. The operation is to improve their functional status on walking and enhance their ability to move [1]. The range of knee flexion is one of the most significant outcome determinants of the Total Knee Arthroplasty (TKA). The maximum flexion smaller than 70 degrees is associated with poor ability to walk, whereas maximum flexion greater than 110 degrees is associated with good ability to walk [2]. In the literature, the association between range of knee flexion and posterior condylar offset (PCO) remains controversial [2-4]. Several studies concluded on a strong correlation [2,3], whereas a study concluded no correlation [4]. This controversy provides a research area needed to be further examined in detail. Moreover, PCO is among the parameters defined as surgically modifiable-factors. Proper assessment in preoperative period and evaluation of this parameter in postoperative period are significant in clinical practice [5].

The PCO was initially defined by Bellemans et al. as "maximal thickness of the posterior condyle projecting posteriorly to the tangent of the posterior cortex of the femoral shaft on a true lateral radiograph" [3]. The measurement of PCO requires an accurate knowledge of the specific magnification factor on that radiograph. The improper standardization of radiographs in terms of magnification factor, leads to inaccurate PCO measurements. The errors are likely to originate from uncertain magnification factors and absence of a reference measurement [6]. In this regard, two novel parameters, known as posterior condylar offset ratio (PCOR) and anterior condylar offset ratio (ACOR), were defined as quantifiable ratios, which do not depend on the magnification factor or reference measurement [6,7]. A. Malviya et al. defines the usage of PCOR instead of PCO as advantage due to no magnification correction issue [5].

PCOR was defined by P. Johal et al. as "the maximal thickness of the posterior condyle projecting posteriorly to a straight line drawn as the extension of the posterior femoral shaft cortex, divided by the maximal thickness of the posterior condyle projecting posterior to a straight line drawn as the extension of the anterior femoral shaft cortex on a true lateral radiograph of the distal quarter of the femur" [8]. C.E.H. Scott et al. studied both PCOR and ACOR. PCOR was defined as PCO divided by femoral diameter [5,7]. ACOR was defined as ACO divided by femoral diameter [7]. The reference values for PCOR and ACOR were introduced as a result of several cohort studies. According to the study conducted by P. Johal et al. on

This work was supported in part by the TUBITAK (The Scientific and Technological Research Council of Turkey).

G. R. Cakmak is with International School of Medicine, Istanbul Medipol University, Istanbul, Turkey. (gulsade.cakmak@std.medipol.edu.tr)
I. E. Hamamci is with International School of Medicine, Istanbul Medipol University, Istanbul, Turkey. (ibrahim.hamamci@std.medipol.edu.tr)
M. K. Yilmaz is with School of Medicine, Istanbul Medipol University, Istanbul, Turkey. (dryilmazkursat@gmail.com)
R. Alhajj is with School of Engineering and Natural Sciences, Istanbul Medipol University, Istanbul, Turkey. (ralhajj@medipol.edu.tr)
I. Azboy is with Department of Orthopedics and Traumatology, Istanbul Medipol University, Istanbul, Turkey. (iazboy@medipol.edu.tr)
M. K. Ozdemir is with School of Engineering and Natural Sciences, Istanbul Medipol University, Istanbul, Turkey. (mkozdemir@medipol.edu.tr)

twenty postoperative radiographs, the mean PCOR was calculated as 0.47 [8].

The measurement/calculation of these parameters, PCO, ACO, PCOR and ACOR, is not commonly performed by the orthopedic surgeons, only for research purposes. The prevalence of obtaining those parameters for TKA-patients is low due to the manual process requiring both time and attention. To extend the usage of these parameters in clinical settings, we developed a software, AutoCOR, which can calculate PCOR and ACOR and visualize the landmarks on plain postoperative lateral knee radiographs within a second. AutoCOR integrates the unsupervised machine learning algorithm (k-means clustering) and image processing techniques (edge detection) for this purpose. The basic mechanism of action is rooted in landmark detection.

When the current literature is examined, the anatomical angle measurement studies were mainly concentrated on landmark detection by deep learning. Several studies regarding the determination of anatomical points on X-ray images by convolutional neural network method exist. One study determines 19 cephalometric landmarks from skeletal X-ray images. The model has two stages: ROI extraction and landmark detection. First, the skeletal X-ray images are divided into patches around the potential anatomical points. These potential/coarse anatomical points are found by image registration method. Intensity-based registration method with mean squared error is applied. The pre-trained ResNet50 model is used for the detection of 19 landmarks on those ROI patches [9]. Another study is performed to determine 3 anatomical points on lower limb X-ray images to measure hip-knee-ankle (HKA) angle. The segmentation by U-Net algorithm is performed in this study. Three areas, ankle, knee and hip joints, are segmented on the X-ray. The center of three areas is found and the HKA angle is measured [10].

In another study, different angles are calculated by detecting 9 anatomical points on lower limb X-rays. The model consists of two stages. In the first stage, first-order detection is applied to find the coarse locations of each point. Then, around those coarse locations, ROI patches are formed. The model is trained by those ROI patches [11].

To sum up the literature review, several studies regarding the landmark detection on lower extremity plain radiographs exist and generally utilized Deep Convolutional Neural Network (CNN). However, no study on autonomous measurement of PCOR and ACOR by using image processing, machine learning or deep learning exits in the literature. This situation highlights the necessity of artificial intelligence in this era.

## II. MATERIALS AND METHODS

This study is approved by the Ethics Review Committee of Istanbul Medipol University.

### 2.1 Dataset

The selection of imaging modality for the measurement process in the project is based on the information obtained from the literature review and current clinical applications. Conventional radiographs are the most commonly used imaging method. Computed tomography (CT) or MRI can be used when necessary, although not frequently. This is due to the patient's exposure to high doses of ionized radiation on CT imaging. Therefore, they are not routinely used methods [12]. Conventional radiography was chosen as the primary modality in the project as a result of its prevalence in the literature and in the clinical settings. Although conventional radiographs are commonly used in preoperative and postoperative periods for TKA patients, disadvantage of this modality is stated as lack of cartilage tissue of knee joint on plain radiographs. This property distorts appropriate preoperative measurement/calculation of offset and offset ratio [8]. In this manner for preoperative measurement, MRI modality appears to give more accurate results, due to its capability to visualize the soft tissue besides the bones.

50 true postoperative lateral knee X-ray images were obtained from the database PUSULA of Istanbul Koşuyolu Medipol Hospital. The images of 32 patients (26 females, 6 males), who underwent TKA in this hospital between 2017 and 2020, were utilized. The images were selected by a trauma and orthopedic surgeon with 14-year of experience. The true radiographs were downloaded in JPEG format. The size of images is 1378 x 672 px.

### 2.2 Image Preprocessing

The original images are cropped between 500 to 900 pixels in width and 224 to 448 pixels in height. Thus, the size of region-of-interest (ROI) patches are determined as 400 x 224 px (Figure-1). ROI patches contain only the knee-joint region with femoral condyles. ROI images are processed further with cv2.bilateralFilter(), cv2.cvtColor() and cv2.threshold() respectively [13]. The aim of preprocessing is to make the femoral component prominent on the ROI patch.

**a) cv2.bilateralFilter():** To apply the threshold method, the image needs to be blurred. There are several blurring methods in the OpenCV library. The selection of the bilateralFilter() method is due to its distinct blurring mechanism. Combined domain and range filtering, provides edge-preservation, noise-removal and smoothing. The d (diameter) is set to 30, sigmaColor and sigmaSpace are set to 100. These parameters are adjusted experimentally to obtain the sharp edges and less noise.

**b) cv2.cvtColor():** Image format is converted from RGB to Grayscale.

**c)cv2.threshold():** The combination of cv2.THRESH_TOZERO and cv2.THRESH_OTSU is used as a thresholding technique.

### 2.3 Image Segmentation by K-means Clustering

The format of thresholded image is converted to RGB and image matrix is reshaped to (-1,3). K-means Clustering, which is an unsupervised machine learning algorithm, is utilized to segment the image in terms of color basis [14]. The main aim is to extract the femoral and tibial components from their surroundings. The criteria-parameter is specified as cv2.TERM_CRITERIA_EPS with 10 iterations and an epsilon value of 1. The number of clusters (k value) is determined as 4 and the technique of cv2.kmeans() is set to cv2.KMEANS_RANDOM_CENTERS.

The number of clusters is determined as 4, and the brightness of the components are the highest when compared with other tissues. The colors of 4 clusters are listed and the brightest one is selected. Brightest cluster is the one which has the biggest decimal code indicating its color. In other words, the decimal value closest to (255,255,255) among these 4 clusters belongs to the femoral and tibial component. The model is mainly built on the basis of this rationale.

The contours around the components are found on the masked image, which contains only the brightest cluster as stated above. The method cv2.findContours() is used to find the edges with cv2.RETR_EXTERNAL technique. This technique is preferred over other ones, as it can detect the largest and comprehensive contours, avoiding subsets. The extraction of the femoral component is mediated by a snippet of code. The algorithm is based on the comparison of contour areas, since the femoral component has the largest area.

### 2.4 Detection of Anterior and Posterior Edge Points of Femoral Component

The extraction of the femoral component allows detection of the edge points of the component on the anterior and posterior side. The coordinates are sorted in ascending and descending order of width (x-axis). The height is sorted in descending order (y-axis). The leftmost and rightmost points are the first element of these sorted lists. Since the coordinates of the edge points are compatible with the ROI patch, they are re-calculated to be visualized on the original image. In this manner, the amount slicing is added to coordinates appropriately. The original image was sliced between 500 to 900 px in width and 224 to 448 px in height. The 500 is added to x values of the edge points, 224 is added to y values of edge points.

### 2.5 Detection of Anterior and Posterior Cortex of Femoral Shaft

The original image is sliced between left edge point to right edge point with 50-pixel extension to both sides in terms of width. It is sliced down to the level of the edge point in terms of height. By this way, a new ROI patch, containing only the femoral shaft, is created. In order to detect the anterior and posterior cortex of the femoral shaft, two lines extending from these cortexes are found. 2 points residing on the anterior cortex line and 2 points residing on the posterior cortex line are selected at the level of 100 px and 140 px in height. The determination of these levels are under the guidance of an orthopedic surgeon. At the level of 100 and 140, the points intersecting with the cortex of the shaft are detected.

To simplify the landmark detection process, the image is processed with Canny Edge Detection, following the bilateral filtering and conversion into grayscale format. Thus, an image containing only the anterior and posterior cortex of the shaft is created. Afterwards, in total 4 points are selected at the 100 and 140 px level. The two corresponding lines are formed by utilizing these landmarks. The image processing steps can be observed below (Fig.1).

### 2.6 Measurement of Femoral Diameter, Anterior and Posterior Condylar Offset

The outcomes of Steps 2.4 and 2.5 provide the anatomical landmarks which are needed to measure Femoral Diameter, ACO and PCO. The measurement of these parameters are mediated by using the np.linalg.norm() function of Numpy Library [15]. The Equation-1 represents the formula to calculate the perpendicular distance between a point and a line. The line equations corresponding to cortexes of the shaft are calculated by using the Line () method of Sympy Library [16].

*(Equation-1)*

$$Line\ passing\ through\ two\ points:$$
$$P_1 = (x_1, y_1) \quad P_2 = (x_2, y_2)$$

$$Point : (x_0, y_0)$$

$$distance = \frac{|(x_2 - x_1)(y_1 - y_0) - (x_1 - x_0)(y_2 - y_1)|}{\sqrt{(x_2 - x_1)^2 + (y_2 - y_1)^2}}$$

ACO is the perpendicular distance between anterior edge point and anterior cortex of the femoral shaft. PCO is the perpendicular distance between posterior edge point of implant and posterior cortex of femoral shaft. The femoral diameter is the perpendicular distance between two cortexes. According to these definitions, the pixel-based measurement of the parameters is completed. The zoom applied to the plain radiographs used in this study varies from 0.2 to 0.4. Therefore, pixel-based measurements cannot be converted into centimeter-based format.

### 2.7 Calculation of Anterior and Posterior Condylar Offset Ratio

The calculation of the ACOR and PCOR is not dependent on the magnification factor. Thus, it can give robust and accurate results. As stated in Step 2.5, linalg.norm() function of Numpy Library and Line() method of SymPy Library are utilized. The important aspect of the proposed model is that both left and right lateral knee X-rays can be given without prior selection of the X-ray type. The model works in the same manner for both left and right knees. The results are analyzed and specification of the limb as left or right knee is done by the software itself. According to this specification, the results are given appropriately under the correct classes.

### 2.8 Analysis of the software AutoCOR

The ground truth ACOR and PCOR values calculated by an orthopedic surgeon were compared with the model outputs. The mean, standard deviation, Bland-Altman Plot are the parameters used to evaluate the performance of the AutoCOR.

### 2.9 Statistical Analysis of Model

The normality test (The Shapiro-Wilk Test) was applied to assess the distribution of the results before correlation test. The Pearson correlation coefficient test was applied between the model output and ground truth values of both ACOR and PCOR.

### III. Results

The model outputs and the manually labeled X-rays which are used as ground truth values can be observed above (Figure-2). The first row contains the model output images. In total 6 anatomical points were selected to measure PCOR and ACOR in an autonomous manner. The proposed model is basically performing a landmark detection. 2 points reside on the anterior cortex and 2 points reside on the posterior cortex of the femoral shaft. These 4 points constitute the tangent lines of cortexes. Remaining 2 points reside on the anterior and posterior edge points of the femoral component. An illustration of usage of these six points and 2 offset ratios can be observed below (Figure-3). As it can be seen from the results, the landmarks are predicted successfully.

### 3.1 Normality Tests and Statistical Analysis

The statistical analysis was performed on Microsoft® Excel® 2016 MSO (version 16.0.4266.1001) and XLSTAT® software (version 2022.1.2.1283, Addinsoft, New York). The ground truth and model outputs were tested in terms of normality. The Shapiro-Wilk Test was used for PCOR and ACOR values. Significance level was adjusted to 0.05 (5%). The P-P and Q-Q plots were generated for the visualization purpose (Fig.4). The minimum, maximum, mean and standard deviation for each class was calculated (Table-1).

Distribution of the data was needed to be evaluated before assessing the correlation between ground truth and model results. The mean (SD) results for PCOR model-data and ground truth-data are 0.984 (0.235) and 0.972 (0.164), respectively. The mean (SD) results for ACOR model-data and ground truth-data are 0.107 (0.092) and 0.107 (0.070), respectively.

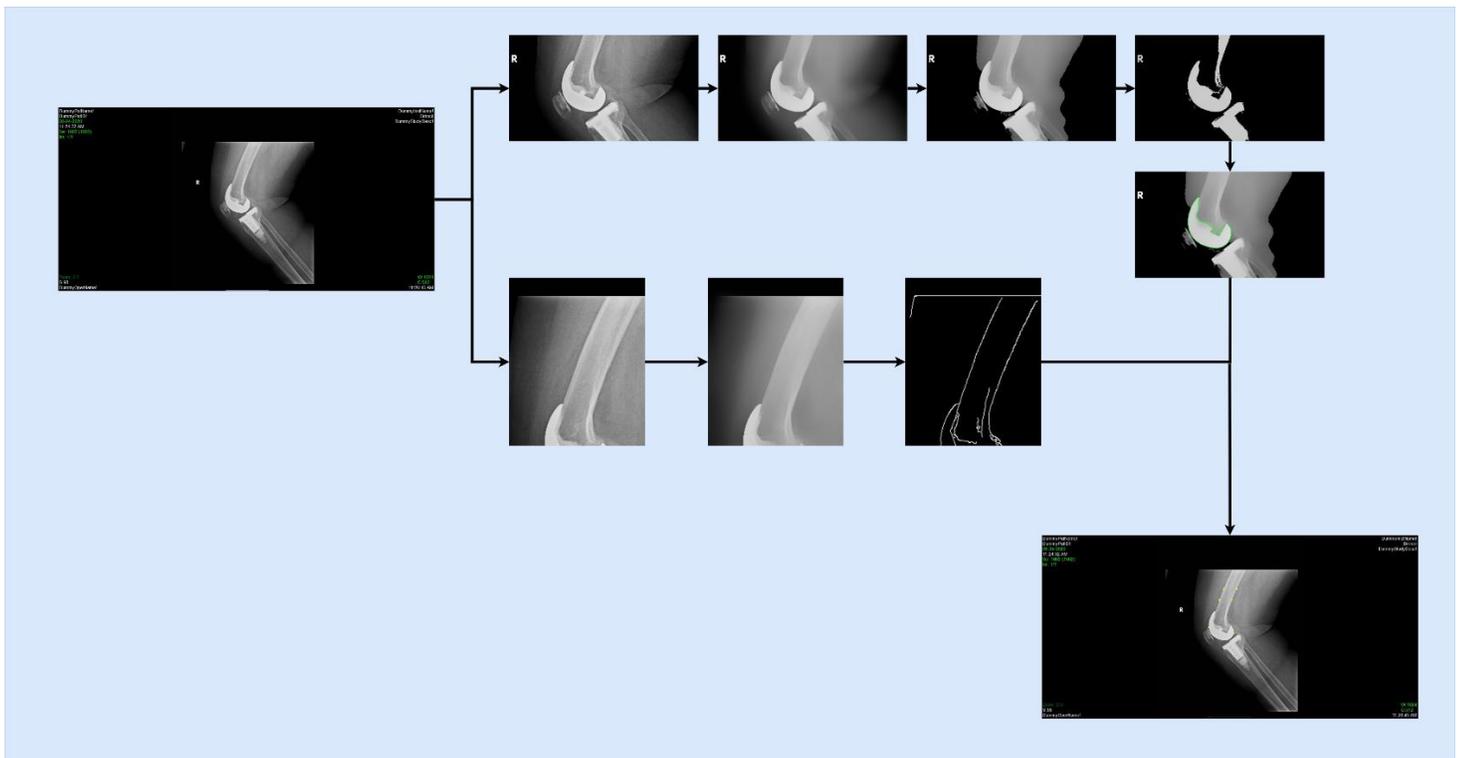

**Figure-1:** Image Processing and K-means Clustering for Landmark Detection

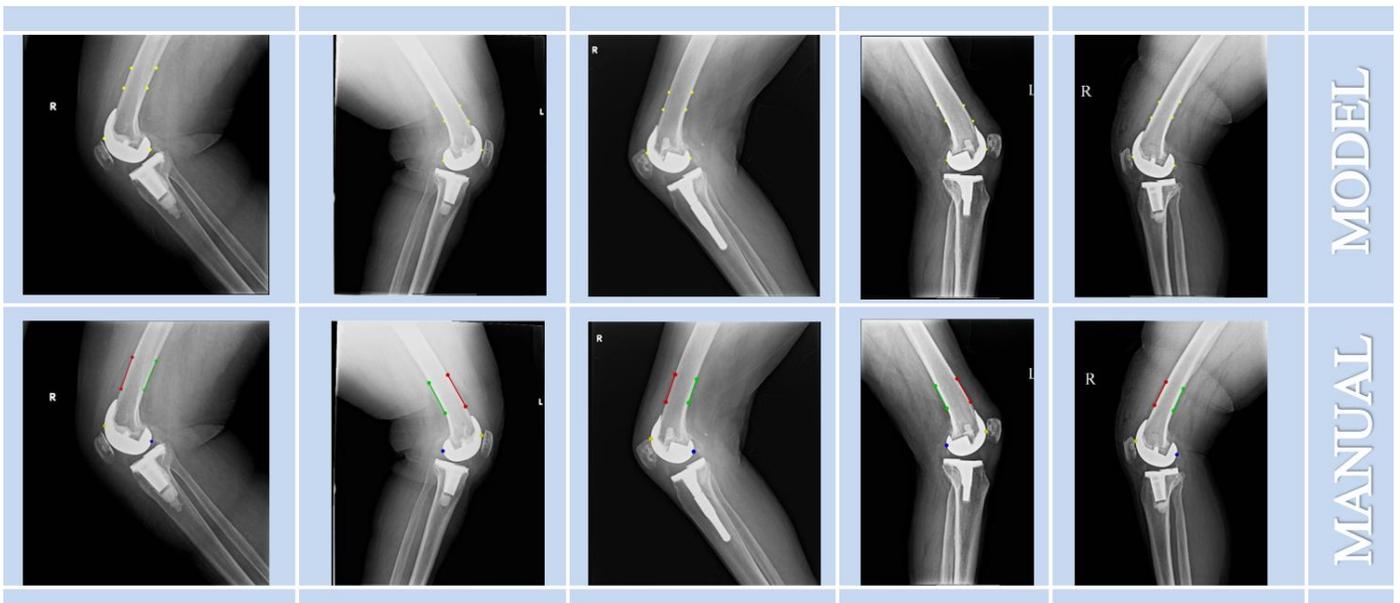

**Figure-2:** The Model Outputs (First Row-Yellow Points); Ground Truth Points (Second Row)

The Shapiro-Wilk Test resulted in "Normal Distribution" for both model and true PCOR data (W=0.982 p=0.635 - W=0.973 p=0.294). The Shapiro-Wilk Test resulted in "Not Normal Distribution" for both model and true ACOR data (W=0.825 p=<0.0001 - W=0.952 p=0.042).

The null hypothesis states a normal distribution, whereas the alternative hypothesis states that the variable from which the sample was extracted, does not follow a normal distribution. The p-value which is greater than the significance level alpha=0.05 indicates that one cannot reject the null hypothesis H0. Thus, indicates a normal distribution. The p-value which is smaller than the significance level alpha=0.05 indicates that one should reject the null hypothesis H0, and accept the alternative hypothesis Ha. In this manner, ACOR data is not distributed normally, and Pearson correlation tests cannot be applied.

### 3.2 Correlation Tests

For PCOR; Pearson Correlation Coefficient Test was applied. The correlation matrix and the p-value were calculated. The correlation between model output and ground truth values is positive and strong (r=0.845). The Pearson Coefficient appears to be significant at the significance level of 0.05 (p<0.0001) (Table-2, 3). Histogram of 2 classes and scatter plot of their combinations are given below (Figure-5).

For ACOR; the normality test indicated a non-normal distribution. Due to lack of normality, to assess the correlation Spearman's correlation coefficient test was applied. Spearman's test does not require normality and it is a non-parametric statistic. The correlation matrix and the p-value were calculated. The correlation between model output and ground truth values is positive and moderate (rs=0.519). The Spearman's Coefficient appears to be significant at the significance level of 0.05 (p=0.0001412) (Table-4, 5). Histogram of 2 classes and scatter plot of their combinations are given below (Figure-5). The p-value indicates the presence of monotonic correlation. The p-value is not equal to zero, referring to presence of a monotonic correlation and accepting the alternative hypothesis.

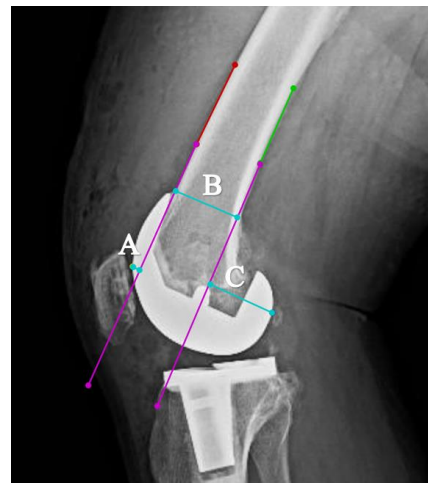

**Figure-3:** The purple lines are the tangent lines of cortexes. The blue lines (A, B, C) represent the anterior condylar offset, femoral diameter and posterior condylar offset, respectively. The ACOR is formulated as A/B and PCOR is formulated as C/B. [7]

**Table-1:** The Statistical Analysis of the Dataset and Model    Outputs (GT: Ground Truth)

| Variable | Observations | Obs. with missing data | Obs. without missing data | Minimum | Maximum | Mean | Std. deviation | W (Shapiro-Wilk Test) | p-value (two-tailed) | Alpha (Shapiro-Wilk Test) |
|---|---|---|---|---|---|---|---|---|---|---|
| PCOR (Model) | 50 | 0 | 50 | 0,508 | 1,576 | 0,984 | 0,235 | 0,982 | 0,635 | 0,050 |
| ACOR (Model) | 50 | 0 | 50 | 0,000 | 0,522 | 0,107 | 0,092 | 0,825 | <0,0001 | 0,050 |
| PCOR (GT) | 50 | 0 | 50 | 0,671 | 1,308 | 0,972 | 0,164 | 0,973 | 0,294 | 0,050 |
| ACOR (GT) | 50 | 0 | 50 | 0,007 | 0,278 | 0,107 | 0,070 | 0,952 | 0,042 | 0,050 |

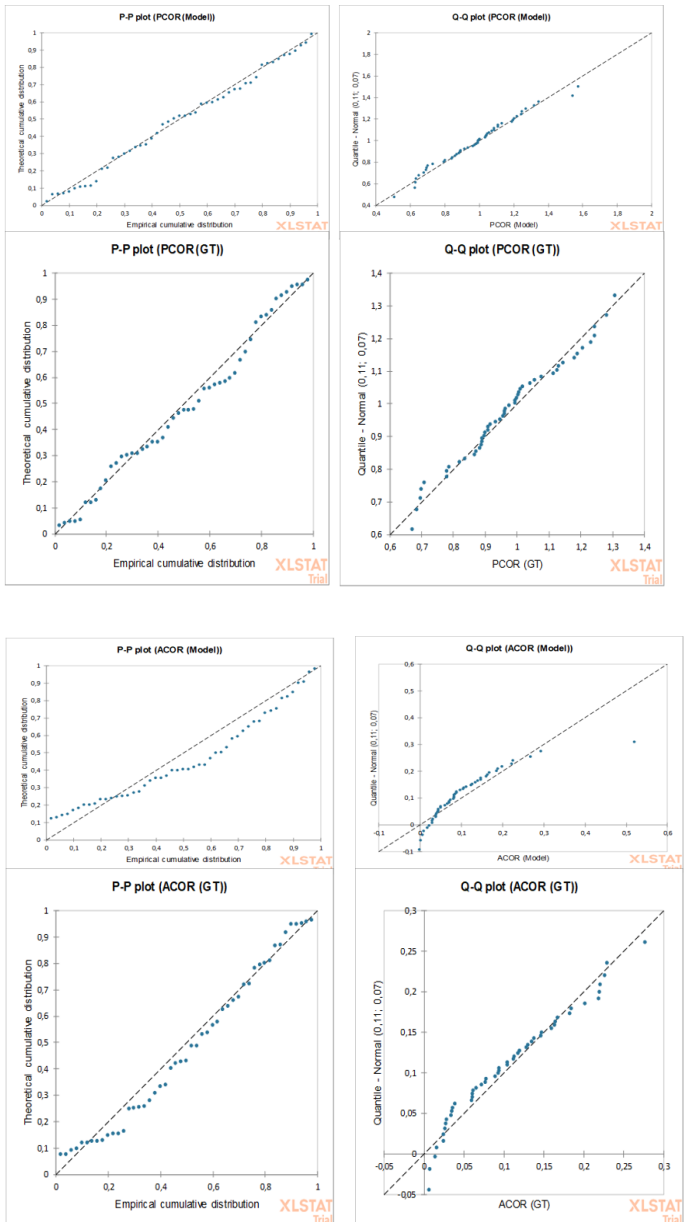

**Figure-4:** The Distribution Scatterplots for PCOR and ACOR Data (GT: Ground Truth)

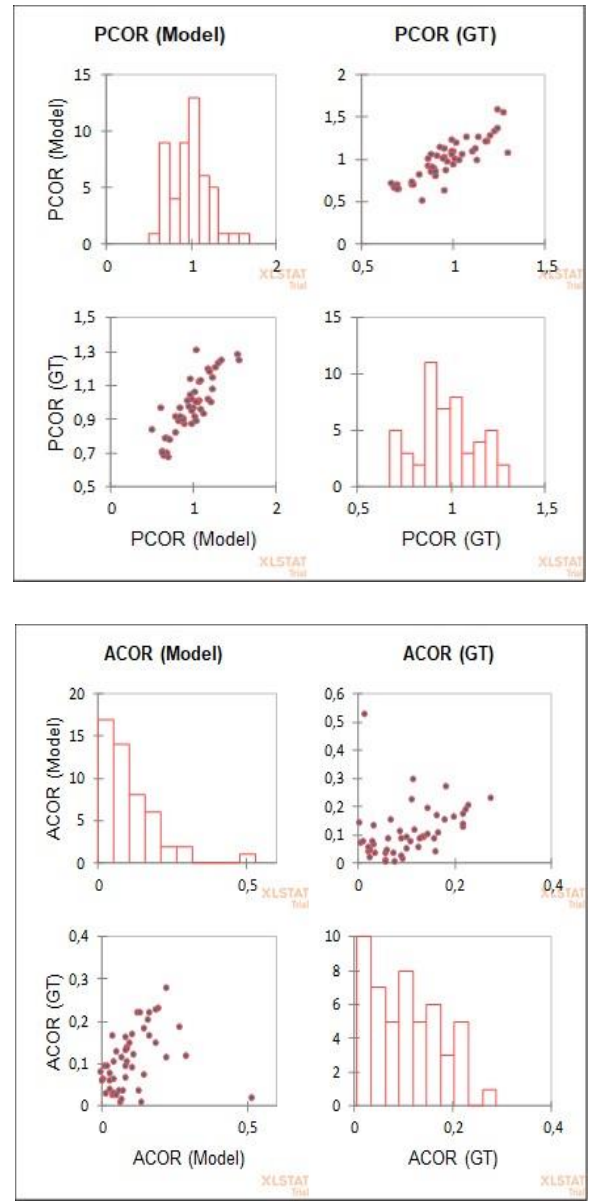

**Figure-5:** Histogram and Scatter Plots for PCOR and ACOR data (GT: Ground Truth)

**Table-2:** Pearson Correlation Matrix

**Table-3:** Pearson p-value Matrix

| Variables | PCOR (Model) | PCOR (GT) |
|---|---|---|
| PCOR (Model) | 1 | 0,845 |
| PCOR (GT) | 0,845 | 1 |
| Values in bold are different from 0 with a significance level alpha=0,05 | | |

| Variables | PCOR (Model) | PCOR (GT) |
|---|---|---|
| PCOR (Model) | 0 | <0,0001 |
| PCOR (GT) | <0,0001 | 0 |

**Table-4:** Spearman's Correlation Matrix (GT: Ground Truth)

**Table-5:** Spearman's p-value Matrix

| Variables | ACOR (Model) | ACOR (GT) |
|---|---|---|
| ACOR (Model) | 1 | 0,519 |
| ACOR (GT) | 0,519 | 1 |
| Values in bold are different from 0 with a significance level alpha=0,05 | | |

| Variables | ACOR (Model) | ACOR (GT) |
|---|---|---|
| ACOR (Model) | 0 | 0,000141918 |
| ACOR (GT) | 0,000141918 | 0 |

### 3.3 Bland-Altman Plots and Histogram of Difference

*1) PCOR:*

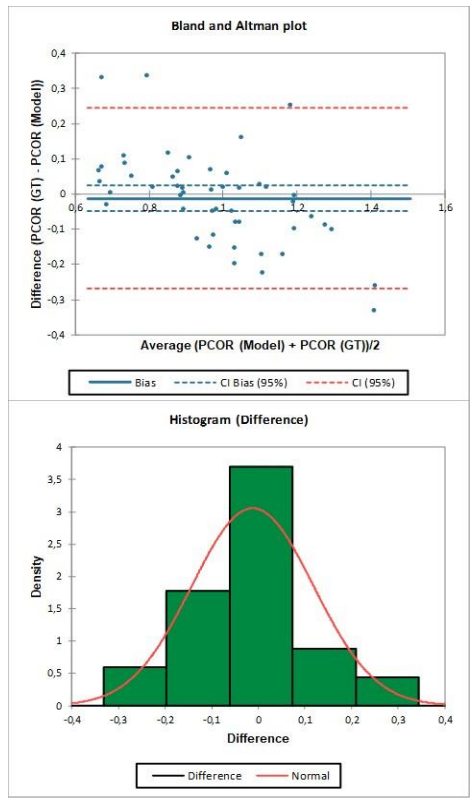

*2) ACOR:*

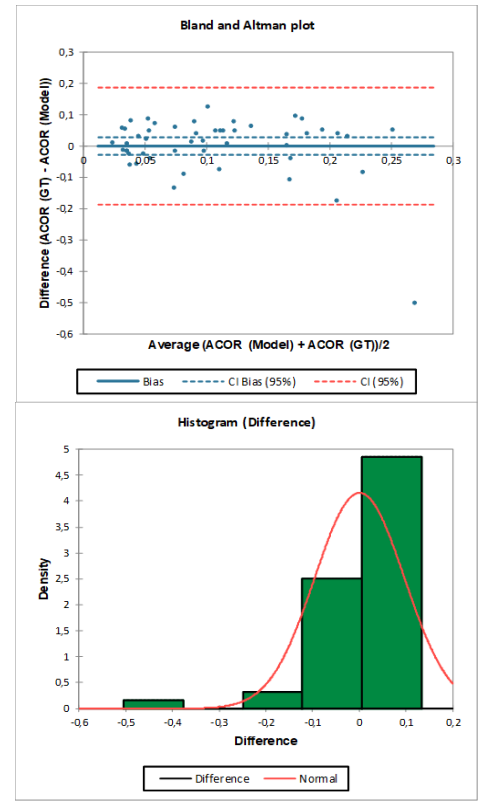

**Figure-6:** Bland-Altman Plot and Histogram PCOR

**Figure-7:** Bland-Altman Plot and Histogram ACOR

## IV. DISCUSSION

The range of knee flexion is a measure of Total Knee Arthroplasty (TKA) outcome. Although the impact of Posterior Condylar Offset (PCO) on the knee flexion is not proved for certain, there are several studies indicating an association. Since a normal kinematic of the knee is not restored by TKA, the evaluation of the current kinematic behavior maintains its importance. The measurement of PCO in the preoperative and postoperative periods provides an opportunity to the physicians to assess the limb-restoration achieved by TKA. The main concern of this measurement is the lack of standard magnification factor on X-rays (110%-130%). The alterations on the magnification factor result in unstable PCO values in different institutions/ hospitals. In the study conducted by P. Johal et al., a novel parameter, Posterior Condylar Offset Ratio (PCOR), was introduced with an advantage of being independent from the magnification factor. C.E.H. Scott et al. studied both PCOR and ACOR.

The proposed model, AutoCOR, is able to calculate the PCOR and ACOR on postoperative lateral knee X-rays by utilizing k-means clustering and image processing techniques. The model, working in the manner of landmark detection, can detect anterior/ posterior edge points of the femoral component and anterior/posterior cortex of the femoral shaft. The model is tested by a dataset containing 50 postoperative lateral knee X-ray from the database PUSULA of Istanbul Medipol Koşuyolu Hospital. The dataset is manually labeled by an orthopedic surgeon to obtain the ground truth values. The PCOR analysis revealed that the mean (standard deviation) of model output (0.984 (SD 0.235)) was close to the ground truth level (0.972 (SD 0.164)). The Shapiro-Wilk normality test, which was applied to determine the distribution of the PCOR values, turned out as "Normal Distribution" (Model: W=0.982 p=0.635 -Ground Truth: W=0.973 p=0.294). Pearson correlation coefficient test produced a strong, positive and significant correlation between model output and ground truth values (r=0.845 p<0.0001).

The ACOR analysis revealed that the mean (standard deviation) of model output (0.107 (SD 0.092)) was close to the ground truth level (0.107 (SD 0.070)). The Shapiro-Wilk normality test, which was applied to determine the distribution of the ACOR values, turned out as "Not Normal Distribution" for both model and true ACOR data (W=0.825 p=<0.0001 - W=0.952 p=0.042). In this manner, ACOR data is not distributed normally, and Pearson correlation tests cannot be applied. Spearman's test revealed "moderate and significant correlation" (rs=0.519 p=0.0001412).

The analysis of the obtained results was mediated by Bland-Altman Plots and Histogram of Difference (Fig.6,7). It can be concluded that the software, AutoCOR, can calculate the PCOR and ACOR successfully, with slight deviation. The software can be improved further and become a clinically-usable tool.

The limitations of the study were the non-normality of the ACOR data. In addition, we are unable to convert pixel-based measurements, PCO, ACO and Femoral Diameter, to centimeter scale. It was due to the presence of different zoom scales on the dataset. In order to extend the scope of the software, the X-ray images should have the same zoom and magnification factors.

## V. ACKNOWLEDGMENTS

This research was supported by the TUBITAK (The Scientific and Technological Research Council of Turkey) as 2209-A Project. We would like to express our very great appreciation to Medipol Koşuyolu Hospital for their support and data provision.